\documentclass[10pt,twocolumn,letterpaper]{article}

\usepackage{iccv}              

\definecolor{iccvblue}{rgb}{0.21,0.49,0.74}
\usepackage[pagebackref,breaklinks,colorlinks,allcolors=iccvblue]{hyperref}
\usepackage{algorithmic}
\usepackage{algorithm}
\usepackage{multirow} 
\usepackage{pifont}
\usepackage{colortbl}
\usepackage{array}
\def\paperID{7029} 
\def\confName{ICCV}
\def\confYear{2025}

\title{Prompt-driven Transferable Adversarial Attack on Person Re-Identification with Attribute-aware Textual Inversion}

\author{Yuan Bian\textsuperscript{1,2}, Min Liu\textsuperscript{1,2}, Yunqi Yi\textsuperscript{1,2}, Xueping Wang\textsuperscript{3}, Shuai Jiang\textsuperscript{1,2}, Yaonan Wang\textsuperscript{1,2}\\
\textsuperscript{1}School of Artificial Intelligence and Robotics, Hunan University\\
\textsuperscript{2}National Engineering Research Center of Robot Visual Perception and Control Technology\\
\textsuperscript{3}College of Information Science and Engineering, Hunan Normal University\\
{\tt\small \{yuanbian, liu\_min, y0512321, wang\_xueping, svyj, yaonan\}$@$hnu.edu.cn}
}

\begin{document}
\maketitle
\renewcommand{\thefootnote}{}  
\renewcommand{\footnotemark}{} 
\begin{abstract}
    Person re-identification (re-id) models are vital in security surveillance systems, requiring transferable adversarial attacks to explore the vulnerabilities of them. 
    Recently, vision-language models (VLM) based attacks have shown superior transferability by attacking generalized image and textual features of VLM, but they lack comprehensive feature disruption due to the overemphasis on discriminative semantics in integral representation.
    In this paper, we introduce the Attribute-aware Prompt Attack (AP-Attack), a novel method that leverages VLM's image-text alignment capability to explicitly disrupt fine-grained semantic features of pedestrian images by destroying attribute-specific textual embeddings. 
    To obtain personalized textual descriptions for individual attributes, textual inversion networks are designed to map pedestrian images to pseudo tokens that represent semantic embeddings, trained in the contrastive learning manner with images and a predefined prompt template that explicitly describes the pedestrian attributes. 
    Inverted benign and adversarial fine-grained textual semantics facilitate attacker in effectively conducting thorough disruptions, enhancing the transferability of adversarial examples. Extensive experiments show that AP-Attack achieves state-of-the-art transferability, significantly outperforming previous methods by 22.9\% on mean Drop Rate in cross-model\&dataset attack scenarios. The codes are at \url{https://github.com/yuanbianGit/AP-Attack}.
\end{abstract}
\begin{figure}[t]
    \centering
    \includegraphics[width=1.0\linewidth]{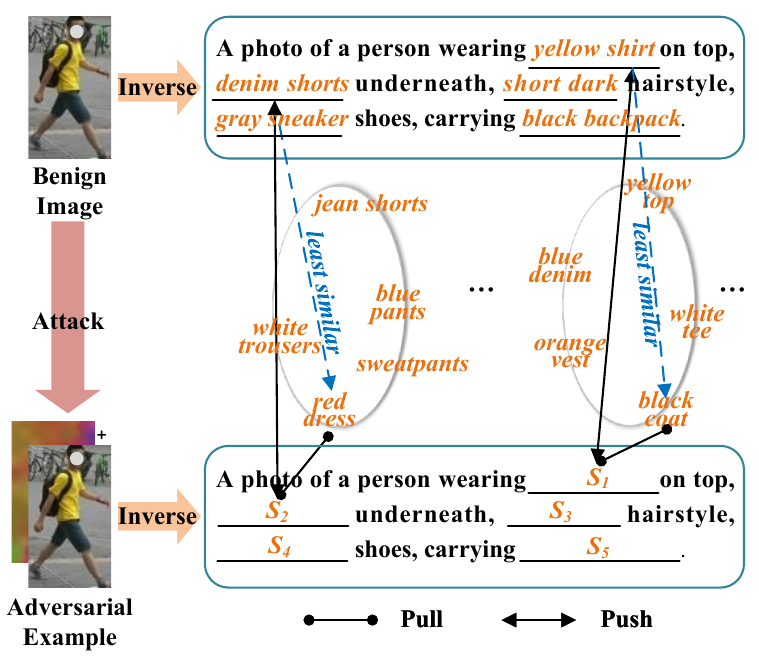}
    \caption{The core idea of our Attribute-aware Prompt Attack. We leverage the image-text alignment capability of vision-language model to invert pedestrian image into pseudo-word tokens $S_*$ that represent attribute-specific semantics, guided by a predefined prompt template explicitly describing person attributes. 
    With inverted semantics, our method enables fine-grained attack by pushing adversarial semantics away from benign ones and pulling them toward the least similar semantics, achieving thorough disruption across all attribute semantic spaces.}
    \label{fig:1}
\end{figure}
\footnote{Corresponding author: Min Liu (liu\_min$@$hnu.edu.cn)}
\vspace{-0.2cm}
\section{Introduction}
\label{sec:intro}
Person re-identification models are widely employed in security-critical surveillance systems, aiming to retrieve the target person \cite{ye2021deep,zheng2016person}. Despite the significant progress made by deep learning-based re-id methods \cite{ahmed2015improved,wang2018cascaded,liu2023weakly,he2021transreid,bian2023occlusion,liu2024A}, they also inherit the vulnerability of deep neural networks, \ie, the addition of imperceptible perturbations on benign images can destroy model performance \cite{goodfellow2014explaining,moosavi2017universal}. The intriguing transferability of adversarial examples (AEs) across different models \cite{szegedy2014intriguing} further exposes real-world surveillance systems to safety threats. To obtain reliable re-id models, it is paramount to test the robustness of re-id models by generating highly transferable AEs.

Cross-model and cross-dataset transferability are crucial for adversarial attacks on black-box re-id models due to the uncertainty architecture of the target model and the significant domain gap between training data and unseen query images \cite{yang2024prompt,salzmann2021learning}. While extensive researches have been conducted to enhance cross-model transferability via input transformation \cite{dong2019evading,xie2019improving}, gradient modification \cite{dong2018boosting,gao2020patch}, model ensembling \cite{li2020learning,liu2016delving}, and intermediate feature attacks \cite{huang2019enhancing,wang2021feature}, few studies have attempted to attack generalized features \cite{zhang2021beyond,li2023cdta} or maximize the fooling gap \cite{naseer2019cross} to improve cross-dataset transferability. These methods primarily depend on the choice of surrogate models, focusing on their architectural similarity to the target model and the generality of the features they extract \cite{ye2024mutual}.

Lately, vision-language models \cite{radford2021learning,jia2021scaling,yang2022unified,yaofilip}, such as CLIP \cite{radford2021learning}, have demonstrated excellence in learning generic representations by training on large-scale Internet image-text pair data and the joint vision-language space of VLM enables zero-shot transfers across downstream tasks with natural language prompts \cite{zhou2022learning,mokady2021clipcap,sanghi2022clip}.
Appreciating the generalization advantage and image-text alignment capability of VLM, some most recent studies \cite{aich2022gama, ye2024mutual, fang2024clip, zhang2022towards} have introduced VLM to facilitate the cross-model and cross-dataset transferability of AEs.
These methods not only delivered optimal loss gradients by incorporating generic image features, but also introduced predefined prompts like `A photo of [CLASS]' \cite{aich2022gama,fang2024clip} or learnable prompts such as `[$V_1$] [$V_2$]\dots [$V_M$] [CLASS]' \cite{yang2024prompt,ye2024mutual}, to guide the implicit semantic destroy in textual cues, thus improving the transferability of AEs.

However, the above VLM-based attack methods solely leveraged global image features or integral textual semantic representation to steer the attacker's learning, which may hinder transferability by overemphasizing discriminative local features and resulting in less comprehensive disruption. To ensure broader and thorough damage to underlying representation, explicitly disrupting fine-grained semantic features is crucial.
Nevertheless, destroying fine-grained semantic features in existing prompt-driven attack methods on the classification task is challenging because the attributes of each category differ, making it impractical to create fine-grained attribute guidance through text prompts.
But it is worth noting that the re-id task is a retrieval task focusing on distinguishing different identities within the pedestrian category, where all images share the same semantic attributes (\eg, gender, clothing, hairstyle).
Therefore, we can leverage VLM's powerful cross-modal comprehension capability to guide the fine-grained image feature disruptions by perturbing semantic text prompts that describe specific attributes of a person, like \cref{fig:1} shows, leading to more thorough disruption and consequently enhancing the transferability of adversarial person images.

Based on the above analysis, we propose a novel Attribute-aware Prompt Attack (AP-Attack) method to achieve transferable fine-grained semantic perturbations on person re-identification. Specifically, vision-language model CLIP \cite{radford2021learning} is adopted in our method, and the adversarial generator is trained to produce delta perturbations.
In pursuit of explicit person attributes information, we construct a personalized prompt template for individual images: \textit{`A photo of a person wearing \underline{$S_1$} on top, \underline{$S_2$} underneath, \underline{$S_3$} hairstyle, \underline{$S_4$} shoes, carrying \underline{$S_5$}.'}, in which pseudo-tokens $S_*$ denotes semantic language description related to each attribute. To obtain these $S_*$, textual inversion technique \cite{galimage}, which learns to capture unique and varied image concepts to a single word embedding, is introduced.
Multiple inversion networks, each corresponding to a specific pedestrian attribute, are designed to generate attribute-aware semantic pseudo-tokens $S_*$, which are then integrated into the predefined template. With composed text prompts and corresponding images, inversion networks are trained in contrastive learning way and subsequently inverse benign and adversarial semantic representations. In this context, prompt-driven semantic attack loss is devised to push the adversarial semantics away from the original ones while pulling them closer to the least similar semantics, guiding the learning of the adversarial generator to destroy fine-grained semantics. By applying this attribute-aware attack across all attribute semantic spaces, our AP-Attack can thoroughly destroy features of pedestrian images, resulting more transferable adversarial examples.

In summary, our main contributions are as follows:
\begin{itemize}
    \item We propose a novel Attribute-aware Prompt Attack method that leverages vision-language model's cross-modal comprehension to perturb fine-grained semantic features of pedestrian images.
    \item To our best knowledge, our method is the first attempt that introduces textual inversion technique to explicitly extract attribute-aware semantic representation to boost the transferability of adversarial examples.
    \item Our AP-Attack achieves state-of-the-art attack transferability across various domains and model architectures, especially surpassing previous approaches by 22.9\% on mean Drop Rate in cross-model\&dataset attack scenarios.
\end{itemize}

\section{Related Works}
\subsection{Adversarial Attack against Re-id}
Re-id models are widely deployed in surveillance systems with stringent security requirements, making their robustness against malicious attacks a critical concern. Unlike classification tasks, re-id is an image retrieval task, and numerous white-box attack methods leveraging adversarial feature similarity metrics have been proposed \cite{bai2020adversarial,zheng2023u,bouniot2020vulnerability}. Given that attackers often need to target unknown models and unseen queries in real-world scenarios, several studies \cite{yang2022towards,ding2021beyond,yang2021learning,wang2020transferable,subramanyam2023meta} have explored cross-model and cross-dataset transferable attacks for black-box re-id systems.
Yang \etal \cite{yang2021learning} and Subramanyam \cite{subramanyam2023meta} improved cross-dataset transferability by utilizing multi-source datasets in meta-learning framework for additive and generative attacks, respectively.
Wang \etal ~\cite{wang2020transferable} presented a Mis-Ranking formulation and multi-stage discriminator network to extract general and transferable features to boost cross-dataset general attack learning. Ding \etal~\cite{ding2021beyond} introduced a model-insensitive regularization technique designed to facilitate universal attacks across diverse CNN architectures. Meanwhile, Yang \etal~\cite{yang2022towards} proposed a combinatorial attack strategy that integrates functional color manipulation and universal additive perturbations to boost the transferability of attacks across both models and datasets.

\begin{figure*}[t]
    \setlength{\belowcaptionskip}{-0.3cm}
    \centering
    \includegraphics[width=1.0\linewidth]{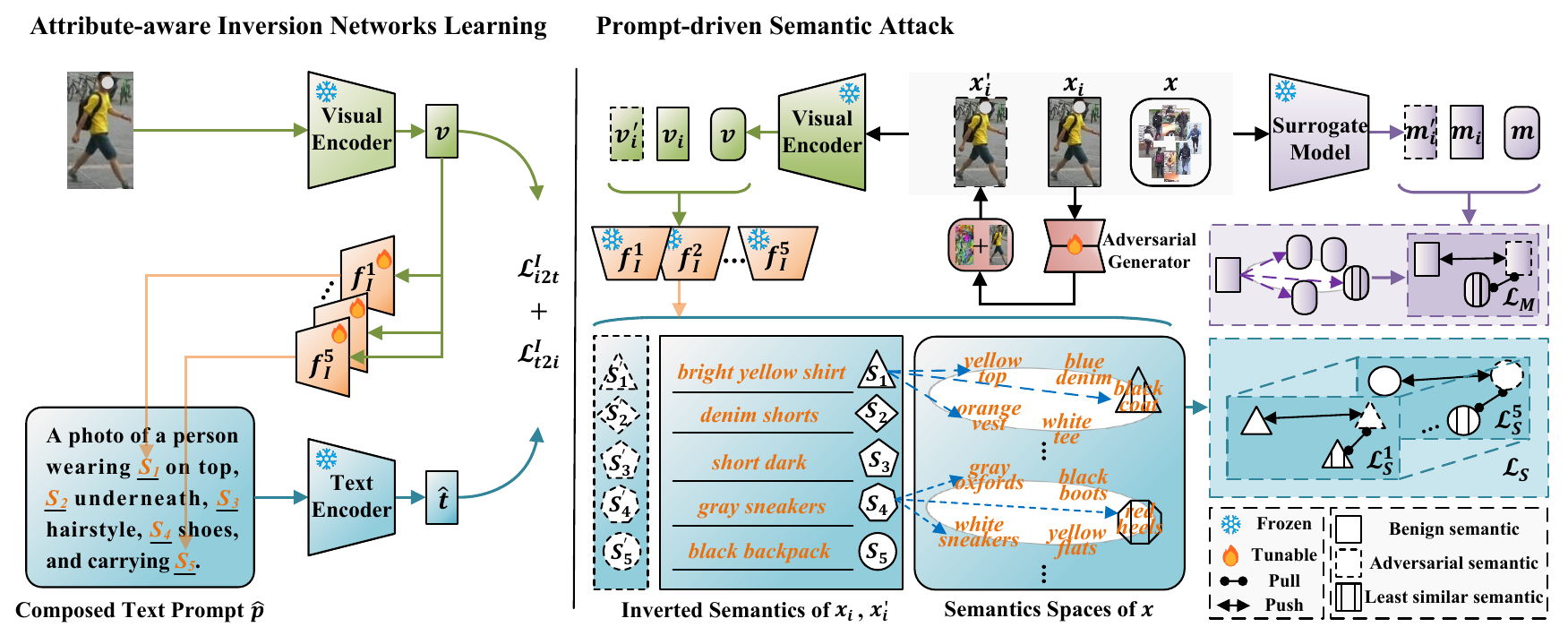}
    \caption{
        The overview of the proposed Attribute-aware Prompt Attack (AP-Attack) method for person re-id. Our AP-Attack follows two stages. First, attribute-aware inversion networks are trained in the contrastive learning manner with benign pedestrian images and composed text prompts. Then, the trained inversion networks are used to guide the prompt-driven semantic attack.
        The generated adversarial examples $x^\prime$, benign images $x$ and batch images $x_b$ are fed into surrogate model and VLM visual encoder to produce inverted semantics and surrogate features of them. 
        The adversarial generator is optimized by pushing the adversarial semantics away from the benign ones and pulling them towards the least similar semantics in semantic spaces of each pedestrian attribute and surrogate feature spaces.}
    \label{fig:2}
\end{figure*}

\subsection{VLM-guided Adversarial Attack}
Vision-language models have garnered significant attention for their ability to learn highly generalizable representations through contrastive pretraining on large-scale image-text pairs \cite{radford2021learning,jia2021scaling,yang2022unified,yaofilip}. 
Given their broad generalization capacity and alignment of visual and language spaces, VLM have become an appealing target for adversarial attacks. Abhishek \etal \cite{aich2022gama} introduced GAMA, the first VLM-based attack targeting multi-object scenes, using text prompts to force adversarial images to align with the least similar text embeddings. Fang \etal \cite{fang2024clip} enhanced the transferability of multi-target adversarial attacks by incorporating VLM textual knowledge to exploit the rich semantic information of target categories. Ye \etal \cite{ye2024mutual} devised an optimization strategy to enhance transferability through iterative attacks on visual inputs while defending text embeddings. Yang \etal \cite{yang2024prompt} proposed PDCL-Attack to facilitate the generalization of classes text feature by prompt learning and formulated a prompt-driven contrastive loss to guide the attack training.
Notably, these VLM-based attack methods are tailored for classification task, leveraging class-specific text labels for guidance. However, re-id aims to distinguish individual identities within the single `person' class, making these approaches unsuitable. The PDCL-Attack method is an exception, as it applies prompt learning to generate prompts for each class. However, its reliance on global semantic features may lead to excessive optimization of highly discriminative features, limiting its effectiveness for thorough destroy for pedestrian images.

In contrast to current person re-id attacks and other VLM-based attacks, our approach seeks to thoroughly undermine fine-grained semantic features by leveraging attribute-aware textual inversion networks, utilizing the image-text comprehension capabilities of VLM.

\section{Method}
In this section, we introduce our AP-Attack method, with an overview provided in \cref{fig:2} and algorithm summary in \cref{alg:algorithm}. The preliminaries of the Contrastive Language-Image Pre-training (CLIP) model and generative adversarial attack definition are presented in \cref{sec:pre}. Details of the attribute-aware textual inversion networks learning are covered in \cref{sec:aati}, followed by the prompt-driven semantic attack process in \cref{sec:psa}.

\subsection{Preliminaries}
\label{sec:pre}
\textbf{Generative Adversarial Attack.} The objective of the proposed AP-Attack is to train the adversarial generator $\mathcal{G}$ to craft perturbations $\mathcal{G}(x)$ for each clean images $x$. The generated perturbations are applied to produce adversarial examples $x^\prime$ by adding perturbations on input images, aiming to deceive the re-id models into retrieving incorrect results. To ensure the perturbations remain subtle and hard to detect, the maximum perturbation magnitude is constrained by a threshold $\epsilon$.
\begin{equation}
    x^\prime = \mathcal{G}(x)+x, \quad\mathrm{s.t.}\|x^\prime-x\|_\infty\leq\epsilon.
    \label{eq:eq0}
\end{equation}
The adversarial generator is initially trained in a white-box setting, where both the data and the surrogate model are known. Once trained, the generator is kept unchanged and employed to generate perturbations for unseen data to attack black-box models.
\begin{algorithm}[t]
    \small
    \caption{Attribute-aware Prompt Attack algorithm}
    \label{alg:algorithm}
    \renewcommand{\algorithmicrequire}{\textbf{Input:}}
    \renewcommand{\algorithmicensure}{\textbf{Output:}}
    \begin{algorithmic}[1] \color{black}
        \REQUIRE Batch images $x$, visual encoder $\mathcal{V}$ and textual encoder $\mathcal{T}$, prompt template $p$, surrogate model $\mathcal{M}$.
        \ENSURE  Inversion networks $f_I$, adversarial generator $\mathcal{G}$
        \STATE Initialize $\mathcal{T}$, $\mathcal{V}$ from pretrained CLIP model and freeze them.
        Initialize $\mathcal{G}$, $f_I$ randomly. Load $\mathcal{M}$ parameters and freeze.
        \WHILE {in \textit{Textual Inversion Learning} process}
        \STATE Extract image features $v$ by $\mathcal{V}$ and inverse $v$ to pseudo-tokens $S_*$ by \cref{eq:3}
        \STATE Integrate $S_*$ into $p$ to form $\hat{p}$ and get text embedding $\hat{t}$
        \STATE Optimize $f_I$ in contrastive learning manner by \cref{eq:6}
        \ENDWHILE
        \STATE Freeze the parameter of $f_I$.
        \WHILE {in \textit{Prompt-driven Semantic Attack} process}
        \STATE Generate adversarial image $x^\prime$ by $\mathcal{G}$
        \STATE Extract image features $m$, $m^\prime$ by $\mathcal{M}$ 
        \STATE Extract image features $v$, $v^\prime$ by $\mathcal{V}$ and inverse them to pseudo-tokens $S_*$, $S^\prime_*$ by \cref{eq:3}
        \STATE Evaluate semantic attack loss by \cref{eq:10} and optimize $\mathcal{G}$ 
        \ENDWHILE
    \end{algorithmic}
\end{algorithm}

\textbf{Contrastive Language-Image Pre-training.} CLIP \cite{radford2021learning} aims to learn highly generalizable representations by utilizing a dataset of 400 million image-text pairs sourced from the internet for language supervision. CLIP is composed of two primary components: a visual encoder $\mathcal{V}(\cdot)$ that processes images $x$ to image features $v$ by $\mathcal{V}(x)$, and a text encoder $\mathcal{T}(\cdot)$ that transforms tokenized text descriptions $p$ to text representation $t$ by $\mathcal{T}(p)$. With image-text batch $\mathcal{S}=\left\{\left(x_n,p_n\right)\right\}_{n=1}^{N}$, the core objective is to align images with their corresponding captions while distinguishing mismatched pairs through contrastive learning by
\begin{equation}
    \mathcal{L}_{i2t} = -\frac{1}{N}\sum_{n=1}^{N}\log\frac{\exp(sim(v_n, t_n)/\tau)}{\sum_{i=1}^N\exp(sim(v_n, t_i)/\tau)},
    \label{eq:1}
\end{equation}
\begin{equation}
    \mathcal{L}_{t2i} = -\frac{1}{N}\sum_{n=1}^{N}\log\frac{\exp(sim(t_n,v_n)/\tau)}{\sum_{i=1}^N\exp(sim(t_n,v_i)/\tau)},
    \label{eq:2}
\end{equation}
where $\tau$ is the temperature scaling factor, and $sim$ represents cosine similarity.

Utilizing the learned joint vision-language feature space, CLIP facilitates zero-shot transfer to a range of downstream tasks by employing natural language prompts to reference the learned visual concepts, \eg, `A photo of a [CLASS]' for classification task. However, for re-id task, where labels are typically index-based, there are no accompanying text labels or descriptions for each identity, making it challenging for CLIP to directly transfer its capabilities to re-id. To tackle this challenge, CLIP-ReID \cite{li2023clip} and PromptSG \cite{yang2024pedestrian} introduced the ID-specific learnable prompts `A photo of a $[X]_1$ $[X]_2$ \dots $[X]_M$ person' and `A photo of a \underline{$S_*$} person', using automated prompt engineering and textual inversion to craft identity representations. Nevertheless, since these prompts lack explicit descriptions of pedestrian attributes, they are not effective for our method, which requires detailed attribute representation for fine-grained attack.
\subsection{Learning Attribute-aware Textual Inversion}
\label{sec:aati}
Textual inversion \cite{galimage} is devised in text-to-image generative task to discover pseudo-words within the text encoder's embedding space that encapsulate both high-level semantics and fine visual details, enabling the generation of new scenes based on user-provided natural language instructions. Textual inversion networks have expanded to composed image retrieval \cite{baldrati2023zero} and person re-id tasks \cite{yang2024pedestrian} to retrieve the target object. These textual inversion methods inverse images into coarse-grained textual semantic representations, where a single word embedding is used to represent the visual information of the entire image. Distinct from them, our methods need to map the fine-grained attribute semantics of pedestrian images.

To explicitly inverse the pedestrian images to attribute-aware semantic representations, we first construct a predefined prompt template \textit{`A photo of a person wearing \underline{$S_1$} on top, \underline{$S_2$} underneath, \underline{$S_3$} hairstyle, \underline{$S_4$} shoes, carrying \underline{$S_5$}.'}, in which five attributes of pedestrian are described and $S_*$ denotes semantic language descriptions related to each attribute. 
Next, several inversion networks that with the same number of preset attributes are designed to map images to pseudo-tokens that represent each attribute semantics. The inverted pseudo-tokens are then composed to predefined text template. Utilizing composed prompts and corresponding images, inversion networks are trained in the contrastive way in the joint vision-language space of CLIP model.

Specifically, five three-layer fully-connected inversion networks, denoted as $f_I$, are constructed. During the training of these inversion networks, both the visual encoder $\mathcal{V}$ and the text encoder $\mathcal{T}$ of pretrained CLIP model are kept frozen to provide the joint vision-language space. The process begins by passing pedestrian images $x$ through the visual encoder $\mathcal{V}$ to extract global visual features $v$. These features are then input into the $i$-th inversion network $f^i_I$, which maps the visual context to attribute-specific semantic pseudo-tokens $S_i$.
\begin{equation}
    S_i=f^i_I(v).
    \label{eq:3}
\end{equation}
All inverted pseudo-tokens $S_*$ are combined to predefined template $p$ to generate a composed language description $\hat{p}$, which is then undergoes a tokenization process and be fed into the text encoder $\mathcal{T}$ to obtain text embedding $\hat{t}$. Using the original image features $v$ and inverted text embedding $\hat{t}$ pairs, inversion networks are trained by the cycle-consistency contrastive loss to ensure learned pseudo-token effectively align with the semantic information of distinct pedestrian attribute. To handle the cases where images are with the same identity that share the same appearance, we follow \cite{yang2024pedestrian, li2023clip} to exploit contrastive loss for re-id as
\begin{equation}
    \mathcal{L}_{i2t}^{I}=\frac1N\sum_{n=1}^N\sum_{c^+\in C(n)}\log\frac{\exp(sim(v_n,\hat{t}_{c^+})/\tau)}{\sum_{i=1}^N\exp(sim(v_n,\hat{t}_i)/\tau)},
    \label{eq:4}
\end{equation}
\begin{equation}
    \mathcal{L}_{t2i}^{I}=\frac1N\sum_{n=1}^N\sum_{c^+\in C(n)}\log\frac{\exp(sim(\hat{t}_n,v_{c^+})/\tau)}{\sum_{i=1}^N\exp(sim(\hat{t}_n,v_i)/\tau)},
    \label{eq:5}
\end{equation}
to ensure learned pseudo-tokens $S_*$ are consistent for the same person, where $C(n)$ represents the corresponding samples sharing the same identity as $v_n$ and $\hat{t}_n$. The total contrastive loss for inversion networks is formulated by
\begin{equation}
    \mathcal{L}_{I} = \mathcal{L}_{t2i}^{I}+ \mathcal{L}_{i2t}^{I}.
    \label{eq:6}
\end{equation}

\subsection{Prompt-driven Semantic Attack}
\label{sec:psa}
Pedestrians are generally recognized as distinct individuals if they differ by even a single semantic information, such as clothing, shoes, or hairstyle, with re-id models relying heavily on these subtle distinctions to accurately identify and differentiate them. In this condition, our method aims to deliberately alter the benign semantic features into other meaningful semantics, thereby misleading the re-id models' recognition. For achieving this, we leverage pretrained inversion networks, which are capable of generating pseudo-tokens that effectively represent visual attribute features within the joint vision-language space, allowing us to produce both adversarial and clean attribute-aware semantic pseudo-tokens to guide the fine-grained semantic attack.

More formally, adversarial images $x^\prime$ are firstly generated by the adversarial generator $\mathcal{G}$ as defined in \cref{eq:eq0}. Then, the benign pseudo-tokens $S_*$ for clean images $x$ and the adversarial pseudo-tokens $S^\prime_*$ for perturbed images $x^\prime$ are obtained through the inversion networks $f_I$. These pseudo-tokens $S_*$ of original batch images form the attribute-specific semantic spaces. In each semantic space, we aim to push the adversarial semantic away from its original images while pulling it closed to its furthest negative semantic by prompt-driven semantic attack loss, which is formulated by
\begin{equation}
    \mathcal{L}^i_{S} = \max\left(0, \|S^\prime_i - S^n_i\|_2 - \|S^\prime_i - S_i\|_2 + \alpha\right),
    \label{eq:7}
\end{equation}
where $S^n_i$ represents the least similar negative semantic in the semantic spaces, $\alpha$ denotes the margin. To boost comprehensive disruption of image features, we apply these constraints across all attribute semantic spaces. The overall prompt-driven fine-grained semantic attack loss is defined as
\begin{equation}
    \mathcal{L}_{S} = \sum_{i=1}^I\mathcal{L}^i_{S},
    \label{eq:8}
\end{equation}
where $I$ is the number of attribute number crafted in prompt template.

Meanwhile, the generated adversarial and clean images are also input into surrogate models $\mathcal{M}$ to get perturbed features $m^\prime$ and clean features $m$. The adversarial attack loss that similar to $\mathcal{L}_{S}$ is conducted by
\begin{equation}
    \mathcal{L}_{M} = \max\left(0, \|m^\prime - m^n\|_2 - \|m^\prime - m\|_2 + \alpha\right),
    \label{eq:9}
\end{equation}
to guide the feature destroy in the surrogate model feature space. Finally, our AP-Attack method optimize the adversarial generator $\mathcal{G}$ by
\begin{equation}
    \mathcal{L}= \mathcal{L}_{M} + \mathcal{L}_{S}.
    \label{eq:10}
\end{equation}
\section{Experiments}
\begin{table*}[t]
    \centering
    \caption{Results of \textbf{cross-dataset} attack: trained on agent model (DukeMTMC) and tested on agent model (MSMT17, Market, CUHK03).}
    \label{tb:tb1}
    \resizebox{0.6\textwidth}{!}{
    \begin{tabular}{c|ccc|c|c}
        \hline \multirow{2}{*}{ Methods } & \multicolumn{3}{c|}{ IDE } & \multirow{2}{*}{ aAP$\downarrow$  } & \multirow{2}{*}{ mDR$\uparrow$ }                                                 \\
        \cline { 2 - 4 }                  & MSMT17                     & Market                              & CUHK03                           &                       &                       \\
        \hline None                       & 41.9                       & 75.5                                & 52.3                             & 56.6                  & -                     \\
        \hline MetaAttack                 & \textbf{3.0}                        & \textbf{4.2}                                 & \textbf{3.8}                              & \textbf{3.7}                   &\textbf{93.5}                  \\
        Mis-Ranking                       & 15.2                       & 26.9                                & 11.1                             & 17.7                  & 68.7                  \\
        MUAP                              & 3.9                        & 19.3                                & 7.6                              & 10.3                  & 81.9                  \\
        \hline GAP                        & 5.9                        & 10.4                                & 5.0                              & 7.1                   & 87.5                  \\
        CDA                               & 7.2                        & 13.3                                & 6.3                              & 8.9                   & 84.2                  \\
        LTP                               & 5.4                        & 9.1                                 & 6.4                              & 7.0                   & 87.7                  \\
        BIA                               & 3.5                        & 14.8                                & 7.0                              & 8.4                   & 85.1                  \\
        PDCL-Attack                       & 4.8                        & 7.4                                 & 7.7                              & 6.6                   & 88.3                  \\
        \hline
        \textbf{AP-Attack(Ours)}                   & 4.2      & 7.6              & 5.3          & 5.7 & 89.9 \\
        \hline
    \end{tabular}
    }
\end{table*}
\begin{table*}[t]
    \centering
    \caption{Results of \textbf{cross-model} attack: trained on surrogate model (DukeMTMC) and tested on victim models (DukeMTMC).}
    \label{tb:tb2}
    \resizebox{0.9\textwidth}{!}{
    \begin{tabular}{c|ccc|cc|ccc|c|c}
        \hline \multirow{2}{*}{ Methods } & \multicolumn{3}{c|}{ Global-based } & \multicolumn{2}{c|}{ Part-based } & \multicolumn{3}{c|}{ Attention-based } & \multirow{2}{*}{ aAP$\downarrow$  } & \multirow{2}{*}{ mDR$\uparrow$ }                                                                                                                            \\
        \cline { 2 - 9 }                  & BOT                                 & LSRO                              & MuDeep                                 & Aligned                             & MGN                              & HACNN                 & Transreid              & PAT                    &                       &                        \\
        \hline None                       & 76.2                                & 55.0                              & 43.0                                   & 69.7                                & 66.2                             & 60.2                  & 79.6                   & 70.6                   & 65.0                  & -                      \\
        \hline MetaAttack                 & 14.9                                & 44.0                              & 31.8                                   & 49.5                                & 57.4                             & 54.6                  & 75.3                   & 64.5                   & 49.0                  & 24.6                   \\
        Mis-Ranking                       & 14.4                                & 6.8                               & 8.0                                    & 16.5                                & 8.4                              & 8.8                   & 34.5                   & 42.9                   & 17.5                  & 73.1                   \\
        MUAP                              & 16.3                                & 9.2                               & 11.1                                   & 23.1                                & 11.4                             & 13.8                  & 34.2                   & 40.4                   & 19.9                  & 69.4                   \\
        \hline

        GAP                               & 12.9                                & 14.6                              & 13.7                                   & 24.5                                & 16.4                             & 16.5                  & 46.7                   & 45.8                   & 23.9                  & 63.3                   \\
        CDA                               & 9.6                                 & 12.5                              & 12.7                                   & 20.8                                & 14.7                             & 15.0                  & 42.3                   & 40.8                   & 21.1                  & 67.6                   \\
        BIA                               & 14.3                                & 33.1                              & 24.5                                   & 44.9                                & 58.0                             & 41.9                  & 71.3                   & 60.8                   & 43.6                  & 32.9                   \\
        LTP                               & 12.3                                & 22.3                              & 23.3                                   & 30.9                                & 37.8                             & 22.5                  & 49.6                   & 45.5                   & 30.5                  & 53.0                   \\
        PDCL-Attack                       & 11.8                                & 11.1                              & 10.5                                   & 22.3                                & 12.6                             & 14.2                  & 37.5                   & 32.0                   & 19.0                  & 70.8                   \\
        \hline
        \textbf{AP-Attack(Ours)}          &\textbf{6.1}             & \textbf{2.2}             & \textbf{6.7}                  & \textbf{6.4}               & \textbf{3.7}            & \textbf{4.7} & \textbf{10.4} & \textbf{15.0} & \textbf{6.9} & \textbf{89.4} \\
        \hline
    \end{tabular}
    }
    \vspace{-0.2cm}
\end{table*}
\begin{table*}[t]
    \centering
    \caption{Results of \textbf{cross-model\&dataset} attack: trained on surrogate model (DukeMTMC) and tested on victim models (Market).}
    \label{tb:tb3}
    \resizebox{0.9\textwidth}{!}{
    \begin{tabular}{c|ccc|cc|ccc|c|c}
        \hline \multirow{2}{*}{ Methods } & \multicolumn{3}{c|}{ Global-based } & \multicolumn{2}{c|}{ Part-based } & \multicolumn{3}{c|}{ Attention-based } & \multirow{2}{*}{ aAP$\downarrow$  } & \multirow{2}{*}{ mDR$\uparrow$ }                                                                                                                              \\
        \cline { 2 - 9 }                  & BOT                                 & LSRO                              & MuDeep                                 & Aligned                             & MGN                              & HACNN                  & Transreid              & PAT                    &                        &                        \\
        \hline None                       & 85.4                                & 77.2                              & 49.9                                   & 79.1                                & 82.1                             & 75.2                   & 86.6                   & 78.4                   & 76.7                   & -                      \\
        \hline MetaAttack                 & 26.3                                & 68.6                              & 37.8                                   & 59.4                                & 73.0                             & 63.9                   & 80.0                   & 67.7                   & 59.6                   & 22.3                   \\
        Mis-Ranking                       & 46.3                                & 36.7                              & 11.9                                   & 47.5                                & 46.7                             & 27.0                   & 65.2                   & 63.4                   & 43.1                   & 43.8                   \\
        MUAP                              & 42.9                                & 35.7                              & 9.7                                    & 48.0                                & 40.6                             & 23.8                   & 58.3                   & 59.7                   & 39.8                   & 48.1                   \\
        \hline GAP                        & 46.1                                & 53.9                              & 19.2                                   & 57.7                                & 60.6                             & 41.8                   & 66.5                   & 67.1                   & 51.6                   & 32.7                   \\
        CDA                               & 46.8                                & 55.9                              & 20.3                                   & 58.5                                & 62.3                             & 46.5                   & 69.0                   & 70.1                   & 53.7                   & 30.0                   \\
        BIA                               & 49.9                                & 60.3                              & 33.9                                   & 61.9                                & 69.8                             & 59.0                   & 78.5                   & 66.1                   & 59.9                   & 21.9                   \\
        LTP                               & 45.3                                & 61.3                              & 32.7                                   & 60.7                                & 67.1                             & 52.6                   & 69.8                   & 68.7                   & 57.3                   & 25.3                   \\
        PDCL-Attack                       & 28.7                                & 36.0                              & 14.4                                   & 40.8                                & 49.7                             & 28.1                   & 61.4                   & 50.8                   & 38.7                   & 49.5                   \\
        \hline
        \textbf{AP-Attack(Ours)}          & \textbf{22.0}              & \textbf{11.1}            & \textbf{6.1}                  & \textbf{24.1}              & \textbf{22.3}           & \textbf{10.3} & \textbf{38.0} & \textbf{35.4} & \textbf{21.2} & \textbf{72.4} \\
        \hline
    \end{tabular}
    }
    \vspace{-0.2cm}
\end{table*}
\subsection{Experimental Setup}
\textbf{Evaluation settings.} To assess the effectiveness of our methods, we comprehensively set cross-model,  cross-dataset and cross-model\&cross-dataset black-box attack scenarios to examine the transferability of generated adversarial examples. 
The cross-model attack setting involves a black-box target model with different architectures from the surrogate model, while sharing the same training dataset. The cross-dataset attack setting, on the other hand, refers to cases where the victim re-id models trained with different dataset but obtain the same network architecture with surrogate models. 
\textbf{Surrogate and victim models.} For the cross-model attack, we choose classical IDE \cite{zheng2016person} model as surrogate model and take BOT \cite{luo2019bag}, LSRO \cite{zheng2017unlabeled}, MuDeep \cite{qian2017multi}, Aligned \cite{zhang2017alignedreid}, MGN \cite{wang2018learning}, HACNN \cite{li2018harmonious}, Transreid \cite{he2021transreid}, PAT \cite{ni2023part} as the victim re-id models. Significantly, these models are built on diverse backbones, including ResNet \cite{he2016deep} (e.g., BOT \cite{luo2019bag}), ViT \cite{dosovitskiy2020image} (e.g., Transreid \cite{he2021transreid}, PAT \cite{ni2023part}), DenseNet \cite{huang2017Densely} (e.g., LSRO \cite{zheng2017unlabeled}), and Inception-v3 \cite{szegedy2016rethinking} (e.g., MuDeep \cite{qian2017multi}). Additionally, these models represent different architecture types, including global-based (e.g., BOT \cite{luo2019bag}), part-based (e.g., MGN \cite{wang2018learning}), and attention-based (e.g., HACNN \cite{li2018harmonious}). \textbf{Training dataset and test dataset.} For the cross-dataset attack, we train our attacker on the surrogate model that pretrained on DukeMTMC \cite{ristani2016performance} dataset and test it on Market \cite{zheng2015scalable}, MSMT \cite{Wei2018Person} and CUHK03 \cite{li2014deepreid} pretrained models.

\textbf{Evaluation metrics.}
We assess the adversarial performance of generated samples against various re-id models using three metrics: mean Average Precision (mAP) \cite{zheng2015scalable}, average mAP (aAP), and mean mAP Drop Rate (mDR) \cite{ding2021beyond}. The aAP is defined as
\begin{equation}
    aAP =\ \frac{\sum_{i=0}^{N}{mAP}_i}{N},
\end{equation}
where ${mAP}_i$ denotes mAP of the $i$-th re-id model. The mDR metric, indicating the success rate of adversarial attacks across multiple models, is calculated as
\begin{equation}
    mDR =\frac{aAP-aAP_{adv}}{aAP},
\end{equation}
where $aAP$ represents the average mAP of the re-id models on the benign images and $aAP_{adv}$ on adversarial examples.

\textbf{Implementation Details.} We adopt the ViT-based CLIP-Reid model \cite{li2023clip} trained on DukeMTMC \cite{ristani2016performance} as the visual and text encoder for CLIP. The adversarial generator follows the Mis-Ranking approach \cite{wang2020transferable}. Optimization is conducted using the Adam optimizer \cite{kingma2014adam} with a learning rate of 2e-4 for both the adversarial generator and the inversion network parameters. All experiments employ $\mathcal{L}_{\infty}$-bounded attacks with $\epsilon=8/255$, setting $\epsilon$ as the maximum change per pixel. The training process of our AP-Attack is implemented in PyTorch and runs on one RTX3090 GPU.

\subsection{Comparison with State-of-the-art Methods}
We evaluate our AP-attack method against state-of-the-art (SOTA) transferable black-box re-id attacks, specifically MUAP \cite{ding2021beyond}, Mis-Ranking \cite{wang2020transferable}, and MetaAttack \cite{yang2022towards}. Notably, MetaAttack also includes color-based perturbations, but for consistency, only its additive perturbation performance is compared. Meanwhile, the state-of-the-art transferable generative attack methods GAP \cite{poursaeed2018generative}, CDA \cite{naseer2019cross}, LTP \cite{salzmann2021learning}, BIA \cite{zhang2022beyond} and PDCL-Attack \cite{yang2025prompt} are incorporated for comprehensive comparisons. It is worth noting that PDCL-Attack \cite{yang2024prompt} is the latest prompt-driven attack method in literature.
All these methods are re-trained with surrogate model IDE \cite{zheng2016person} on DukeMTMC \cite{ristani2016performance} for fair comparisons. 

\textbf{Comparisons on cross-dataset attack.} The results of cross-dataset attack are shown in \cref{tb:tb1}, from which can be seen that our method gets 5.7\% aAP and 89.9\% mDR. Our AP-Attack surpasses the SOTA generative attack method PDCL-Attack by 0.9\% and 1.6\% on aAP and mDR, respectively. 
Comparing to SOTA re-id attack method MetaAttack that incorporates multi-datasets in meta-learning scheme, our method get comparable performances with only one dataset for training.

\textbf{Comparisons on cross-model attack.} Experimental results in \cref{tb:tb2} show that our method achieves the best performances of 6.9\% aAP and 89.4\% mDR score on cross-model scenarios, significantly outperforming the SOTA methods by 10.6\% and 16.3\% in terms of aAP and mDR. 

\textbf{Comparisons on cross-model\&dataset attack.} For the majority of realistic and complex cross-model\&dataset attack results in \cref{tb:tb3}, our AP-Attack method exceeds the SOTA method PDCL-Attack by 17.5\% on aAP accuracy and 22.9\% on mDR, which further highlights the superiority and effectiveness of our method. 

Notably, our method outperforms SOTA prompt-driven attack method PDCL-Attack in all attack settings. The advantage of our method on re-id can be attributed to two main factors. First, our method achieves fine-grained, thorough feature disruption, while PDCL-Attack lacks of comprehensive destroy by perturbing only on global features.  Second, the learned prompts in PDCL-Attack are specifically designed for different IDs within the same class `person' in re-id attack, which differs from prompt learning in classification task where prompts describe different categories, likely leads to prompts that are more stylized rather than universal. In contrast, our approach utilizes the predefined prompt template, guiding the inversion network to convert more specific and generalizable semantic information, resulting in more broadly applicable and transferable results.

\subsection{Ablation Studies}

\begin{figure}[t]
    \centering
    \includegraphics[width=1.0\linewidth]{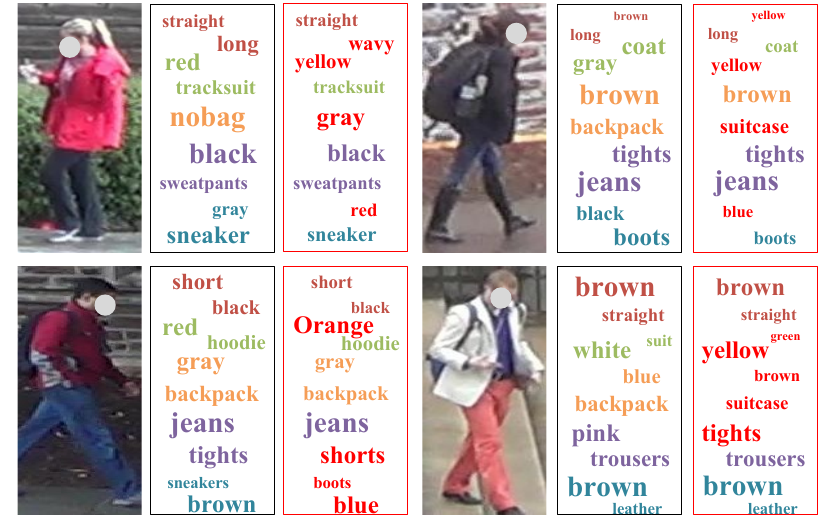}
    \setlength{\abovecaptionskip}{-0.1cm}
    \setlength{\belowcaptionskip}{-0.4cm}
    \caption{Word cloud visualization of the learned attribute-specific pseudo-tokens, where semantic words for benign images are in black boxes and  AEs' are in red boxes. Distinct colors of the word represent different attributes and attributes highlighted in red illustrate the disrupted semantics of AEs. Font size indicates the similarity to tokens, with larger fonts represent greater similarity.
    }
    \label{fig:3}
\end{figure}

\textbf{The effectiveness of textual inversion.} In order to verify the effectiveness of our textual inversion networks, we attempt to interpret the learned pseudo-tokens to meaningful word. We first established an attribute-specific semantic vocabulary \footnote{Details of produced semantic vocabulary shows in supplemental files.} by chatGPT, with each attribute corresponding to a distinct, meaningful set of semantic words. These attribute vocabularies capture both color and descriptive features. 
Then, we calculate the similarity between each word in the vocabulary and the pseudo-tokens, selecting the two words with the highest similarity scores for display, as shown in \cref{fig:3}. From the figure, it can be seen that the semantic words similar to the learned pseudo-tokens can correctly describe the pedestrian image information, indicating the effectiveness of our inversion networks.

\begin{table*}[t]
    \centering
    \caption{Attack effectiveness against defense methods.}
    \label{tb:tb5}
    \resizebox{0.65\textwidth}{!}{
        \begin{tabular}{c|ccc|c|c}
            \hline Method     & Adv.ResNet & Randomization & JPEG(60\%) &$\mathrm{aAP} \downarrow$ & $\mathrm{mDR} \uparrow$ \\
            \hline None       & 69.6       & 84.6          & 83.8       & 80.0  & -     \\
            \hline MetaAttack & 67.1       & 67.8          & 57.9       & 64.3  & 19.7  \\
            Mis-Ranking       & 56.1       & 43.3          & 51.2       & 50.2  & 37.3  \\
            MUAP              & 53.6       & 48.5          & 57.4       & 53.2  & 33.5  \\
            \hline Ours       & 39.0       & 26.6          & 27.8       & \textbf{31.1}  & \textbf{61.1}  \\
            \hline
        \end{tabular}
    }
\end{table*}

\textbf{The effectiveness of fine-grained attack.} To illustrate the superiority of our AP-Attack for guiding fine-grained semantic attack, we compare the results of cross-model\&dataset attack using different adversarial triplet losses that incorporate various features. Specifically, we use the loss constraint based solely on the image features from the surrogate model as the baseline, and compare with the results obtained by adding different CLIP feature losses, including global visual features, integral text embeddings, and fine-grained semantic embeddings. As shown in \cref{tb:tb4}, the results incorporating CLIP feature constraints significantly outperform the baseline. Moreover, the inclusion of text embeddings yields better results than using image features, suggesting that text-based features offer greater universality. Most importantly, our fine-grained semantic embedding constraints achieve the best performance, demonstrating the effectiveness of our method. 

\begin{figure}[t]
    \centering
    \includegraphics[width=0.8\linewidth]{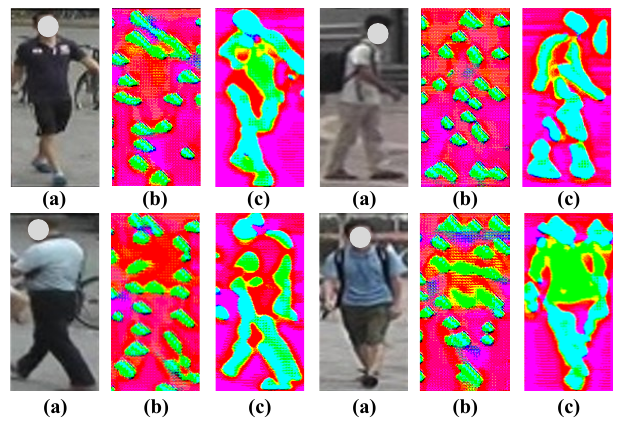}
    \caption{Visualization of perturbations under different feature constraints: (a) shows the original image, (b) depicts the perturbation when incorporating global image features from the CLIP model, and (c) presents the perturbation under our AP-attack with fine-grained semantic feature constraints.}
    \label{fig:4}
\end{figure}

To visually demonstrate that our method performs fine-grained attacks on each attribute, we compared the perturbation images under different feature constraints. As shown in \cref{fig:4}, compared to the incorporating global CLIP feature constraints, the perturbations generated by our method cover a larger area, closely resembling the full pedestrian posture in the image. This indicates that our method produces perturbations that attempt to disrupt all semantic features of the pedestrian in a fine-grained manner. Meanwhile, as can be seen from the attribute word cloud of AEs in Fig. \ref{fig:3}, our approach is destructive to fine-grained semantics, and can destroy most of the attribute semantics.

\subsection{Attack Effectiveness against Defense Method}

 We conduct evaluations against three defense strategies, including adversarially trained models (Adv. Res \cite{bouniot2020vulnerability}), input preprocessing techniques (JPEG compression \cite{das2017keeping}), and denoising-based methods (Randomization \cite{xie2018mitigating}). For JPEG, a compression rate of 60\% is applied, and the victim model is BOT(Market). \cref{tb:tb5} shows that our method consistently achieves superior attack effectiveness across these defenses, achieves an mDR of 61.1\%.
\begin{table}[t]
    \centering
    \caption{Results of cross-model\&dataset attack using different adversarial triplet losses that incorporate various features.}
    \label{tb:tb4}
    \resizebox{0.48\textwidth}{!}{
    \begin{tabular}{c|c|c}
        \hline                           & aAP$\downarrow$ & mDR$\uparrow$ \\
        \hline baseline                  & 45.8            & 40.3          \\
        baseline+global visual feature           & 27.9            & 63.6          \\
        baseline+integral text embedding         & 25.9            & 66.2          \\
        baseline+fine-grained semantic embedding & \textbf{21.2}   & \textbf{72.4} \\
        \hline
    \end{tabular}
    }
    \vspace{-0.2cm}
\end{table}

\subsection{Transferability to Diverse Types Models} 
\textcolor{black}{
    To further evaluate our AP-Attack's generalizability across diverse types of re-id models, we test it against three distinct re-id models: the self-supervised PASS (Market) \cite{zhu2022pass}, auxiliary-feature-enhanced PGFA (Occluded-Duke) \cite{miao2019pose}, and CLIP-ReID (Market) \cite{li2023clip} based on CLIP. Tab. \ref{tb:tb6} reveals substantial performance degradation across all tested architectures, confirming the efficacy of our method across diverse model paradigms.}

\begin{table}[t]
    \caption{\textcolor{black}{Comparisons on self-supervised, auxiliary feature and CLIP-based re-id models.}}
    \label{tb:tb6}
    \centering
    \resizebox{0.47\textwidth}{!}{
    \begin{tabular}{c|cc|cc|cc}
        \hline
        \multirow{2}{*}{Method} & \multicolumn{2}{c|}{PASS} & \multicolumn{2}{c}{PGFA} & \multicolumn{2}{|c}{CLIP-ReID} \\ 
        \cline{2-7}
         & mAP & Rank-1              & mAP & Rank-1       & mAP & Rank-1          \\ 
        \hline
        None                    & 92.2 & 96.3               & 37.3 & 51.4          & 89.6 & 95.5         \\ 
        Ours                   & 12.1 & 14.3               & 4.2  & 5.7         & 32.7  & 40.8          \\ 
        \hline
    \end{tabular}
    }
  \end{table}
\section{Conclusion}
In this paper, we propose a novel Attribute-aware Prompt Attack methods to enhance the transferability of adversarial attacks on person re-id task. Our AP-Attack method leverages the image-text alignment capability of VLM and introduces the attribute-specific inversion networks to map the image feature to attribute semantic textual embeddings. And it attempts to thoroughly destroy the pedestrian features by perturbing fine-grained attribute semantics across all attribute feature spaces. Extensive experimental results validate the superiority of our approach in cross-dataset and cross-model black-box attack scenarios, achieving substantial performance gains over the latest SOTA methods. We believe that our work offers a meaningful contribution to adversarial attack research and holds promise for strengthening the security of machine learning systems in real-world.

\noindent\textbf{Acknowledgments.} This work was supported in part by the National Natural Science Foundation of China under Grant 62425305, U22B2050 and 62221002, in part by the Science and Technology Innovation Program of Hunan Province under Grant 2023RC1048, in part by the Hunan Provincial Natural Science Foundation of China under Grant 2024JJ3013. 
{
    \small
    \bibliographystyle{ieeenat_fullname}
    \bibliography{main}
}

\end{document}